\documentclass[runningheads]{llncs}
\usepackage{multirow}
\usepackage{graphicx}
\usepackage{enumitem}
\usepackage{varwidth}
\usepackage{tasks}
\usepackage{xcolor}
\usepackage{hyperref}

\makeatletter
\newcommand\subsubsubsection{\@startsection{paragraph}{4}{\z@}{-2.5ex\@plus -1ex \@minus -.25ex}{1.25ex \@plus .25ex}{\normalfont\normalsize\bfseries}}
\newcommand\subsubsubsubsection{\@startsection{subparagraph}{5}{\z@}{-2.5ex\@plus -1ex \@minus -.25ex}{1.25ex \@plus .25ex}{\normalfont\normalsize\bfseries}}
\makeatother
\usepackage{amsmath}

\begin{document}
%
% \title{Contribution Title\thanks{Supported by organization x.}}
% \title{Line and Word Segmentation for Bangla
% Handwritten Text Recognition}

\title{\texttt{BN-DRISHTI: Bangla Document Recognition through Instance-level Segmentation of Handwritten Text Images}}

\titlerunning{BN-DRISHTI}

\author{Sheikh Mohammad Jubaer\inst{1} \and
Nazifa Tabassum\inst{1} \and \\
Md. Ataur Rahman\inst{1} \and 
Mohammad Khairul Islam\inst{2}}
\authorrunning{S.M. Jubaer et al.}
\institute{Premier University, Dept. of CSE, Chittagong, Bangladesh\\
\email{\{jubaer, nazifa\}.puc@gmail.com; ataur.cse@puc.ac.bd} \and
University of Chittagong, Dept. of CSE, Chittagong, Bangladesh\\
\email{mkislam@cu.ac.bd}
\vspace{-2mm}}

\maketitle         % typeset the header of the contribution

\begin{abstract}
Handwriting recognition remains challenging for some of the most spoken languages, like Bangla, due to the complexity of line and word segmentation brought by the curvilinear nature of writing and lack of quality datasets. This paper solves the segmentation problem by introducing a state-of-the-art method (\texttt{BN-DRISHTI}\footnote{\textbf{Code and Demo:} \url{https://github.com/crusnic-corp/BN-DRISHTI}}) that combines a deep learning-based object detection framework (YOLO) with Hough and Affine transformation for skew correction. However, training deep learning models requires a massive amount of data. Thus, we also present an extended version of the BN-HTRd dataset comprising 786 full-page handwritten Bangla document images, line and word-level annotation for segmentation, and corresponding ground truths for word recognition. Evaluation on the test portion of our dataset resulted in an F-score of $99.97\%$ for line and $98\%$ for word segmentation. For comparative analysis, we used three external Bangla handwritten datasets, namely BanglaWriting, WBSUBNdb\_text, and ICDAR 2013, where our system outperformed by a significant margin, further justifying the performance of our approach on completely unseen samples.

% Handwriting recognition remains challenging for some of the most spoken languages, like Bangla, due to the complexity of line and word segmentation brought by the curvilinear nature of writing and lack of quality datasets. This paper solves the segmentation problem by introducing a state-of-the-art method (\texttt{BN-DRISHTI}) that combines a deep learning-based object detection framework (YOLO) with Hough and Affine transformation for skew correction. However, training deep learning models requires a massive amount of data. Thus, we also present an extended version of the BN-HTRd dataset comprising 786 full-page handwritten Bangla document images, line and word-level annotation for segmentation, and corresponding ground truths for word recognition. Evaluation on the test portion of our dataset resulted in an F-score of $99.97\%$ for line and $98\%$ for word segmentation. For comparative analysis, we used three external Bangla handwritten datasets, namely BanglaWriting, WBSUBNdb\_text, and ICDAR 2013, where our system outperformed by a significant margin. A manual evaluation over the unannotated portion of our dataset was also carried out in a random setting that gave near-perfect results, further justifying the performance of our system on completely unseen samples.

% \vspace{-2mm}
\end{abstract}

\section{Introduction}
% In government or non-government sectors, especially in Banks, Courts, and Industries where the medium of communication is Bangla, primarily the handwritten data in the documents are also preserved in Bangla; consequently, many decisions have to be made on those data. Despite the vast community of Bangla-speaking people worldwide and Bangla being one of the most recognized languages, not so much emphasis has been given to the Bangla handwritten text recognition task. Therefore, to automate or semi-automate those process of decisions making by implementing Artificial Intelligence (AI) or Machine Learning (ML), the handwritten documents need to be converted into digital images, which means there involve the task of end-to-end handwritten text recognition from the copy of Bangla scripts. 

Line and word segmentation are one of the most fundamental parts of handwritten document image recognition. As the field of deep learning is maturing at an unprecedented speed, the choice for solving this sort of task employing off-the-shelf deep learning frameworks is getting popular nowadays for its efficiency. However, few attempts have been made to utilize this approach for Bangla handwritten recognition task due to the scarcity of datasets in this domain. Our previous endeavors involved an initial dataset-making process named BN-HTRd (v1.0), comprising of Bangla handwritten document images and only line-level annotations and ground truths for word recognition. However, that dataset was incomplete due to the missing word-level annotation. Therefore, to have a more comprehensive and useable handwritten recognition dataset, we have extended the BN-HTRd (v4.0) dataset\footnote{\textbf{Extended Dataset:} \url{https://data.mendeley.com/datasets/743k6dm543}} by integrating word-level annotations and necessary improvements in the ground truths for the word recognition task.

As segmentation plays a vital role in recognizing handwritten documents, another pivotal \emph{contribution} of this paper is the conglomeration of a state-of-the-art method for segmenting lines and words from transcribed images. Our approach treats the segmentation task as an object detection problem by identifying the distinct instances of similar objects (i.e., lines, words) and demarcating their boundaries. Thus in a way, we are performing \texttt{instance-level segmentation} as it is particularly useful when homogeneous objects are required to be considered separately. To do so, we partially rely on the YOLO (You Only Look Once) framework. However, the success of our method is more than just the training of the YOLO algorithm. In order to get the perfect words segmented from possibly complex curvilinear text lines, we had to improvise our approach to retrieve the main handwritten text lines correctly by removing other unnecessary elements. For that, we used a combination of the Hough and Affine transform methods. The Hough transform predicts the skew angles of the main handwritten text lines, and the Affine transform rotates them according to the expected gradients, making them straight horizontally. Therefore, the word segmentation approach provides much better results compared to the segmentation on skewed lines. Thus, the main contributions of this paper are threefold:

\begin{enumerate}
  \item   Introducing a straightforward \texttt{novel hybrid approach}, for instance-level handwritten document segmentation into corresponding lines and words.
  
  \item	Achieved \textit{state-of-the-art} (\texttt{SOTA}) scores on three different prominent Bangla handwriting datasets for line/word segmentation tasks.
  
  \item	Set a new \texttt{benchmark} for the BN-HTRd dataset. Also, \texttt{extended}\footnote{\textbf{Changes:} \url{https://data.mendeley.com/v1/datasets/compare/743k6dm543/4/1}} it to be one of the largest and the most comprehensive Bangla handwritten document image segmentation and recognition dataset by adding \texttt{200k+} annotations.
  \vspace{-2mm}
\end{enumerate}

\section{Related Work}
\vspace{-2mm}
\textbf{CMATERdb} \cite{sarkar2012cmaterdb1} is one of the oldest character-level datasets consisting of 150 Bangla handwritten document images distributed among two versions. Another prominent character-level dataset having 2000 handwritten samples named \textbf{BanglaLekha-Isolated} \cite{biswas2017banglalekha} contains 166105 handwritten characters written by an age group of 6 to 28. \textbf{Ekush} \cite{rabby2019ekush}, which is a multipurpose dataset, contains 367,018 isolated handwritten characters written by 3086 individual writers. The authors also benchmarked the dataset using a multilayer CNN model (\textbf{EkushNet}) for character classification, achieving an accuracy of $97.73\%$ on their dataset while scoring $95.01\%$ in the external \textbf{CMATERdb} dataset.

A paragraph-level dataset that resembles our dataset in terms of word-level annotation is the \textbf{BanglaWriting} \cite{mridha2021banglawriting} dataset, which includes single-page handwriting comprising 32,787 characters, 21,234 words, and 5,470 unique words produced by 260 writers of different ages and personalities. Another paragraph-level unannotated dataset \textbf{WBSUBNdb\_text} \cite{halder2018content}, consisting of 1383 handwritten Bangla scripts having around 100k words, was collected from 190 transcribers for the writer identification task. While in terms of a document-level dataset, mostly resembling our own, \textbf{ICDAR 2013} \cite{stamatopoulos2013icdar} handwriting segmentation contests dataset comes with 2649 lines and  23525 word-level annotations for 50 handwritten document images on Bangla.

\raggedbottom
Segmenting handwritten document images in terms of lines and words is the most crucial part when it comes to end-to-end handwritten document image recognition. In \textit{\textbf{Projection-based}} methods \cite{fernandez2014graph}\cite{nicolaou2009handwritten}\cite{mullick2015efficient}\cite{boukharouba2017new}, the handwritten lines are obtained by computing the average distance between the peaks of the projected histogram. A method based on the skew normalization process is proposed in \cite{bal2018improved}. \textbf{\textit{Hough-based}} methods \cite{fernandez2014graph} represent geometric shapes such as straight lines, circles, and ellipses in terms of parameters to determine geometric locations that suggest the existence of the desired shape. The author of \cite{boukharouba2017new} presented a skew correction technique for handwritten Arabic document images using their optimized randomized Hough transform, followed by resolving the primary line for segmentation. For layout analysis, \textit{\textbf{Morphology-based}} approaches \cite{fernandez2014graph}\cite{boudraa2017improved} have been used along with piece-wise painting (PPA) algorithms \cite{alaei2011new}, to segment script independent handwritten text lines. In contrast, \textit{\textbf{Graph-based}} approaches \cite{fernandez2014graph}\cite{surinta2014path}\cite{kumar2011segmentation} compactly represent the image structure by keeping the relevant information on the arrangement of text lines. \textbf{\textit{Learning-based}} techniques recently became popular for segmenting handwritten text instances. The authors of \cite{renton2017handwritten}\cite{vo2016dense}\cite{renton2018fully}\cite{barakat2018text} used a Fully Convolutional Network (FCN) for this purpose.
% While \cite{renton2017handwritten}\cite{renton2018fully} trained their network by defining the text lines with X-height labeling, in \cite{renton2017handwritten}, the network is seven-layer architecture, where the first six layers combine convolutions and dilations, and the last layer stands for predictions.Unlike other FCN methods, in \cite{renton2018fully}, they used two distinct layers (7 and 11). However, in \cite{barakat2018text}, authors applied some pre-processing techniques, such as an adaptive binarization method, to annotate line masks.
A model based on the modified multidimensional long short-term memory recurrent neural networks (\textbf{MDLSTM RNNs}) was proposed in \cite{bluche2016joint}. An unsupervised \textbf{\textit{clustering}} approach \cite{rahman2023bn} was utilized for line segmentation which achieved an F-score of $81.57\%$ on the BN-HTRd dataset.

A series of consistent recent works on \textbf{Bangla handwriting segmentation} \cite{rakshit2018line}\cite{agarwal2022word}\cite{rakshit2023generalized} is carried out by a common research team that also developed the WBSUBNdb\_text dataset. Their technique predominantly relies on the projection profile method and connected component analysis. They initially worked on a tri-level (line/word/character) segmentation \cite{rakshit2018line} while their latest works are focused solely on word \cite{agarwal2022word} and line segmentation \cite{rakshit2023generalized}. Moreover, in \cite{rakshit2023generalized}, the method serves the line segmentation on multi-script handwritten documents while the other two research only work for the Bangla scripts.

Our work can be categorized as a \textbf{Hybrid Approach} for segmenting lines and words. Our supervised models employ YOLO deep learning framework to predict lines and words from handwritten document images. We used the Hough Line Transform to measure the segmented line's skew angle, then corrected it with Affine Transform. These combinations were never used in the literature for Bangla handwritten recognition tasks.

\section{Dataset}
\label{sec_3}
\vspace{-2mm}
Data annotation is one of the most crucial parts of the dataset curation process where supervised learning is concerned. As a primary text source, we considered the BBC Bangla News platform since it does not require any restrictions and has an open access policy. Hence, we downloaded various categories of news content as files in TEXT and PDF format, renamed files according to the sequence of 1 to 237, and put them in separate folders. We distributed those 237 folders among 237 writers of different ages, disciplines, and genders. They were instructed to write down the text file's contents in their natural writing style and to take pictures of the pages afterward. This resulted in 1,591 handwritten images in total. Due to the complexity of the task, we were only able to recruit a total of 75 individuals to annotate lines of assigned handwritten images using an annotation tool called \texttt{LabelImg}. As a result, we were only able to annotate a maximum of 150 folders. The resultant annotation produced YOLO and PASCAL VOC formatted ground truth for line segmentation. These 150 folders of handwritten images and their line annotations were included in the first version of the BN-HTRd dataset \cite{rahman2023bn}. For the purpose of word segmentation, we have extended the dataset (v4.0) by adding bounding-box annotations of individual words for all the annotated lines. We also organized each word of the text file into separate rows in Excel in order to create the ground truth Unicode representation of the corresponding word's images for recognition purposes in the future.

We used this extended BN-HTRd dataset containing annotations in 150 folders to develop and test our system. It contains a total of 786 handwritten images comprising 14,383 lines and 1,08,181 words. The rest of the unannotated 87 folders were automatically annotated using our system, resulting in an additional 14,836 lines and 1,06,135 words, which we denoted as Automatic Annotations. For the purpose of experimental evaluation, we split the 150 folders into two subsets and took one image from each of the folders for either validation or testing (resulting in 75 images for each subset). The rest of the 636 images were used for training purposes. Table \ref{drishti-tab1} below shows this subdivision.

\vspace{-4mm}
\begin{table}
\centering
\caption{Distribution of extended BN-HTRd (v4.0) dataset for experimentation.\protect\footnotemark}\label{drishti-tab1}
% \begin{tabular}{|l|l|r|r|r|r|}
\begin{tabular}{|p{0.17\linewidth}|p{0.25\linewidth}|p{0.12\linewidth}|p{0.12\linewidth}|p{0.12\linewidth}|p{0.12\linewidth}|}
\hline
\textbf{Type} &   \textbf{Purpose} &  \textbf{Train} & \textbf{Valid} & \textbf{Test} & \textbf{Total}\\
\hline
\hline
Doc. Images & Line Segmentation & 636 & 75 & 75 & 786\\
\hline
Line Images & Word Segmentation & 11,471 & 1,515 & 1,397 & 14,383\\
\hline
Word Images & Word Recognition & 86,055 & 11,712 & 10,414 & 1,08,181\\
\hline
\end{tabular}
\vspace{-8mm}
\end{table}

\footnotetext{\textbf{Splitted Dataset:} \url{https://doi.org/10.57967/hf/0546}}

\vspace{-2mm}

\section{Proposed Methodology}
\vspace{-1mm}

We have broken down our overall system architecture in Fig. \ref{drishti-fig1}, which consists of six parts. Those six parts cover the overall process of how our system functions. Before dissecting those parts in detail in the later sections (\ref{sec_4_1} - \ref{sec_4_5}), we will provide a brief overview in the following:

\begin{figure}[h]
\includegraphics[width=1\textwidth]{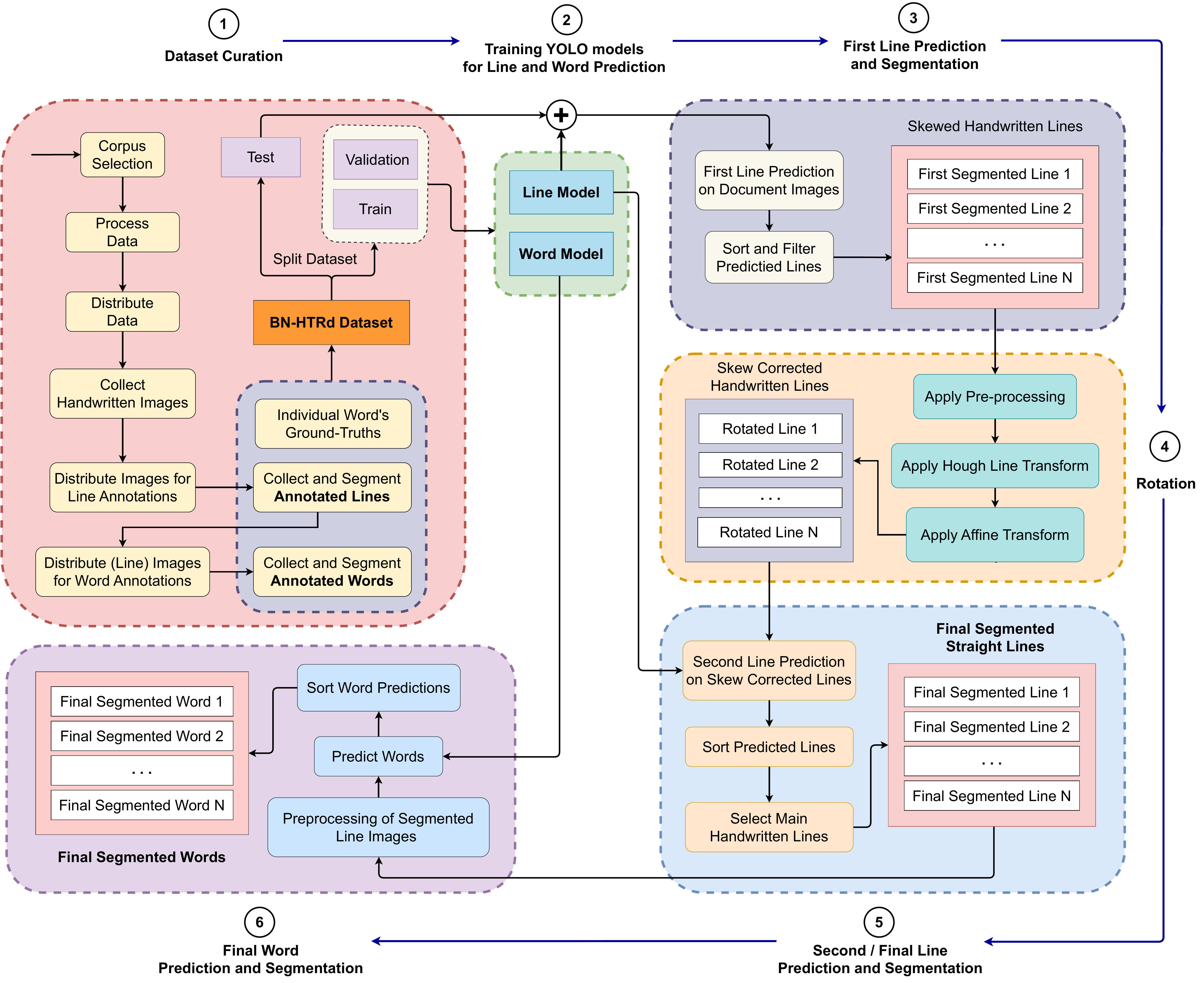}
\vspace{-6mm}
\caption{Overall System Architecture for BN-DRISHTI.} 
\vspace{-4mm}
\label{drishti-fig1}
\end{figure}

\begin{itemize}
\setlength\itemsep{2mm}
    \item Our efforts in making and extending the BN-HTRd dataset involved various development processes such as distributing the data to the writers, manual annotations, and making it compatible with supervised learning methods such as ours (details in section \ref{sec_3}).
    
    \item Although training the models is a crucial part of any supervised system, it was not enough in our case despite YOLO being one of the best frameworks. It was predicting redundant lines, which we had to eliminate in order to get better segmentation scores (details in section \ref{sec_4_1} and \ref{sec_4_2}).
    
    \item As we were also getting some unnecessary lines along with the target line, a better line segmentation method is essential to segment the words correctly. To remove them, we rotated the curvilinear lines using the Hough-Affine transformation and corrected their skewness (details in section \ref{sec_4_3}).
    
    \item We applied the final/second YOLO line prediction on the skew-corrected lines, followed by some post-processing in order to extract the main handwritten line (details in section \ref{sec_4_4}).
    
    \item Finally, word prediction and segmentation are performed on skew-corrected final segmented line images using the word model (details in section \ref{sec_4_5}).
\end{itemize}

\subsection{Training Models}
\label{sec_4_1}
YOLOv5x (XLarge) model architecture having a default SGD optimizer was used to train both our Line and Word models for 300 epochs. We used document images with line annotations to train the initial line segmentation model. In contrast, line images and their word annotations were used to train the Word model. The training was done using an NVIDIA RTX 3060 Laptop GPU containing 6 GB GDDR6 memory and 3840 CUDA cores.

\subsection{First-Line Prediction and Segmentation}
\label{sec_4_2}
% The line detection is done on the document images without any image pre-processing or resizing and some output samples are given in Fig. \ref{drishti-fig2}a. For each document image, YOLO generates a TEXT file where each predicted line is represented by: \textit{$<$$class\_id$, x, y, width, height, confidence$>$} without any particular order. We set the confidence as $0.3$ while predicting since some lines with a few words, especially with one word were missing out as the line model was not considering them as lines. As a result, we had some unnecessary line predictions along with the correct ones having confidence less than $ 0.5$. Hence, after sorting the predicted output based on the y-axis attribute, we also filtered those unnecessary lines, resulting in the filtered first-line prediction of the document images (Fig. \ref{drishti-fig2}b). Afterward, we extract the filtered predicted lines using their YOLO attributes: \textit{$<$ x, y, width, height $>$}. Fig. \ref{drishti-fig2}c illustrates first-line detection and filtering, along with the corresponding first-line segmentation.

\begin{figure}
% \vspace{-4mm}
% \includegraphics[width=1\textwidth]{images/4.png}
\includegraphics[width=1\textwidth]{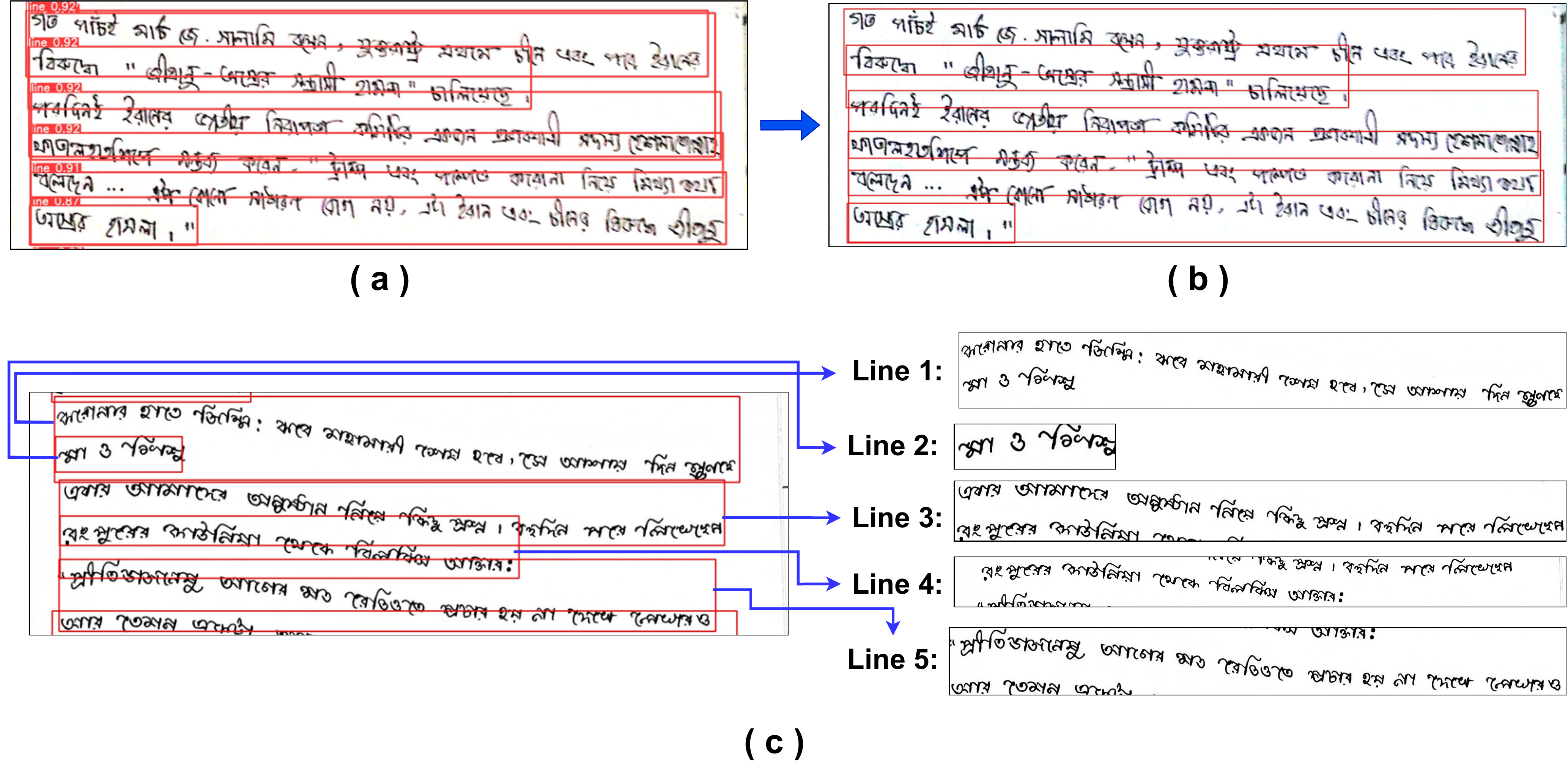}
\vspace{-4mm}
\caption{Representation of First-line prediction and segmentation, where a) sample image with first-line prediction containing multiple unnecessary predictions, b) filtered first-line prediction, and c) another sample image with filtered first-line prediction and segmentation for curvilinear handwriting.}
\label{drishti-fig2}
\vspace{-4mm}
\end{figure}

The line detection is performed on document images without pre-processing or resizing; some output samples are shown in Fig. \ref{drishti-fig2}a. YOLO generates a TEXT file for each document image, representing each predicted line as \textit{$<$$class\_id$, x, y, width, height, confidence$>$} without particular order. The confidence threshold during prediction is set to $0.3$ to include lines with few words or a single word that was initially missing. However, this approach resulted in both unnecessary line predictions and correct ones with confidence below $0.5$. To address this, the output is sorted based on the y-axis attribute, and unnecessary bounding boxes having unusual heights but lower confidence that encompasses or overlaps with one or more boxes are filtered out, resulting in filtered first-line predictions (Fig. \ref{drishti-fig2}b). The filtered predicted lines are then extracted using their YOLO attributes: \textit{$<$ x, y, width, height $>$}. Fig. \ref{drishti-fig2}c illustrates the process of first-line detection, filtering, and corresponding segmentation.

\vspace{-4mm}
\subsection{Rotation (skew estimation and correction)}
\label{sec_4_3}
\begin{figure}[h]
\centering
% \vspace{-6mm}
\includegraphics[width=0.9\textwidth]{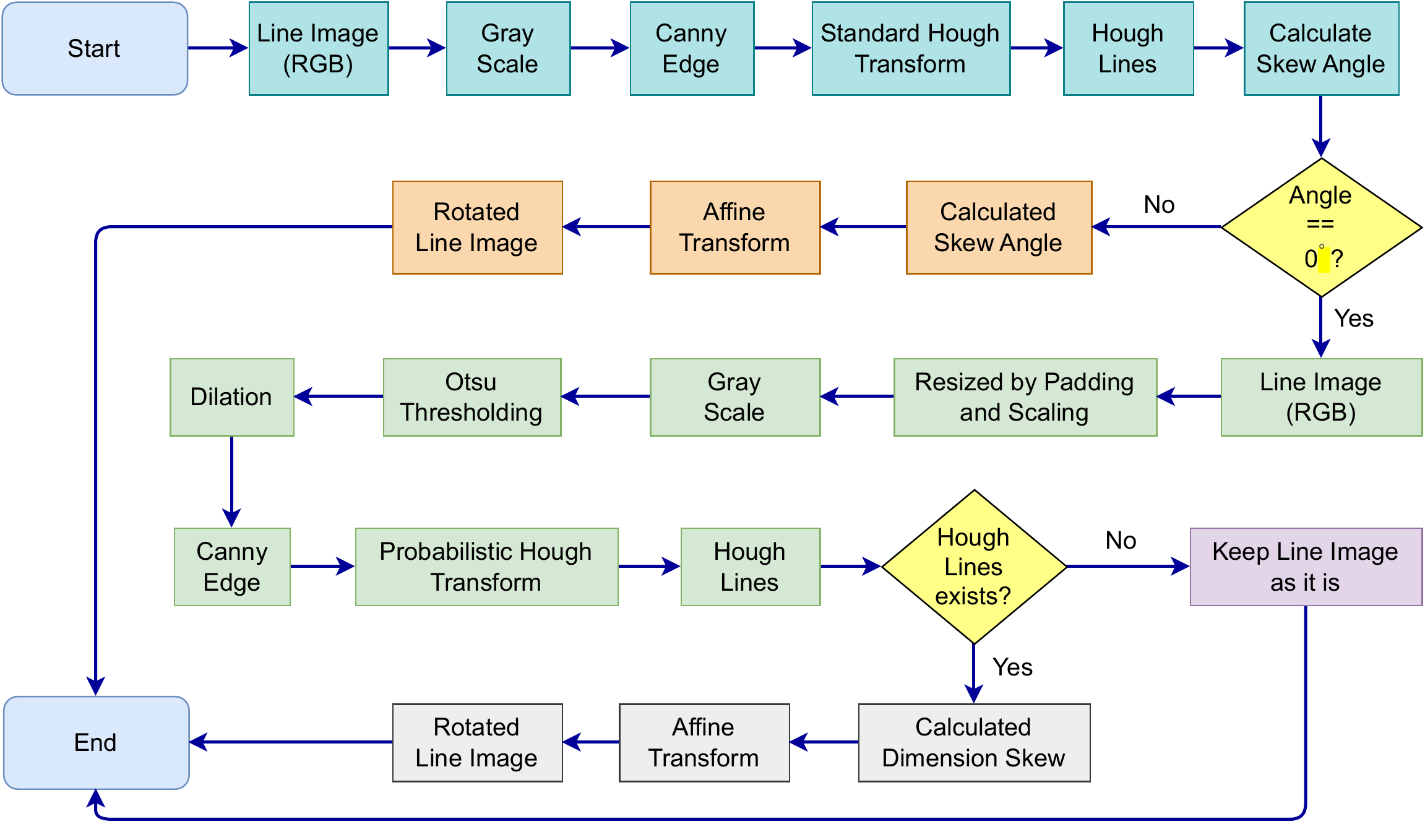}
\vspace{-2mm}
\caption{Flowchart of skew estimation and correction over the first predicted lines.}
\label{drishti-fig3}
\vspace{-6mm}
\end{figure}

After analyzing our first segmented line images, we found out that, with the main handwritten line, we are also getting some unwanted lines at the top or bottom due to the skewness of the lines and the rectangular shape of the predicted bounding box. Therefore, the skew correction over the first line prediction is important in order to retrieve the main handwritten line. We denoted this process as \textit{Rotation}, which is performed by applying the Hough line and Affine transform. We have represented the overall rotation process in Fig. \ref{drishti-fig3}.

\subsubsection{Skew Estimation:}

\begin{figure}
\centering
% \vspace{-4mm}
\includegraphics[width=1.0\textwidth]{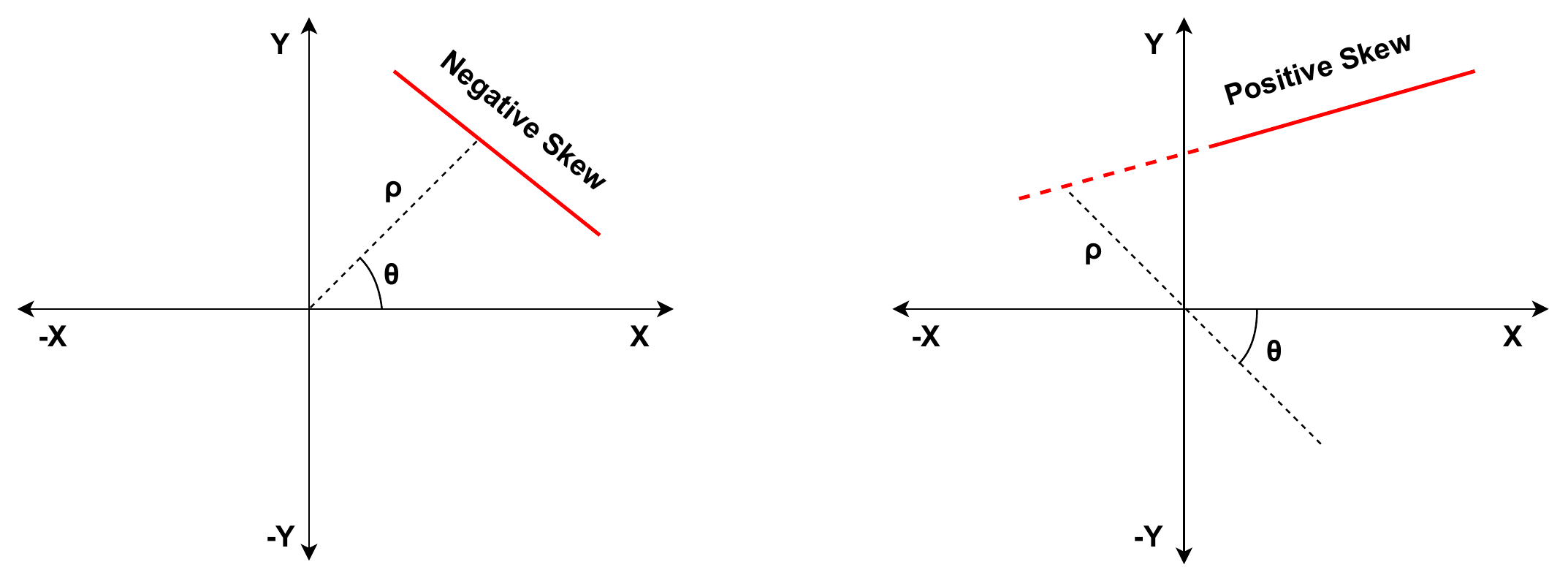}
\vspace{-6mm}
\caption{Representation of Hough lines using equation $\rho$ = x*$\cos\theta$ + y*$\sin\theta$; where $\theta$ is the angle of the detected line and $\rho$ is the distance from x-axis.} 
\label{drishti-fig4}
\vspace{-4mm}
\end{figure}

We categorized handwritten lines' skew into two types: Positive and Negative (shown in Fig. \ref{drishti-fig4}). The skew angle estimation is performed in two phases:
\begin{enumerate}
    \item Line Skew (LSkew) Estimation: where we applied the Standard Hough Transform (SHT).
    \item Dimension Skew (DSkew) Estimation: where we applied the Probabilistic Hough Transform (PHT).
\end{enumerate}

\begin{figure}
\centering
\vspace{-4mm}
\includegraphics[width=1.0\textwidth]{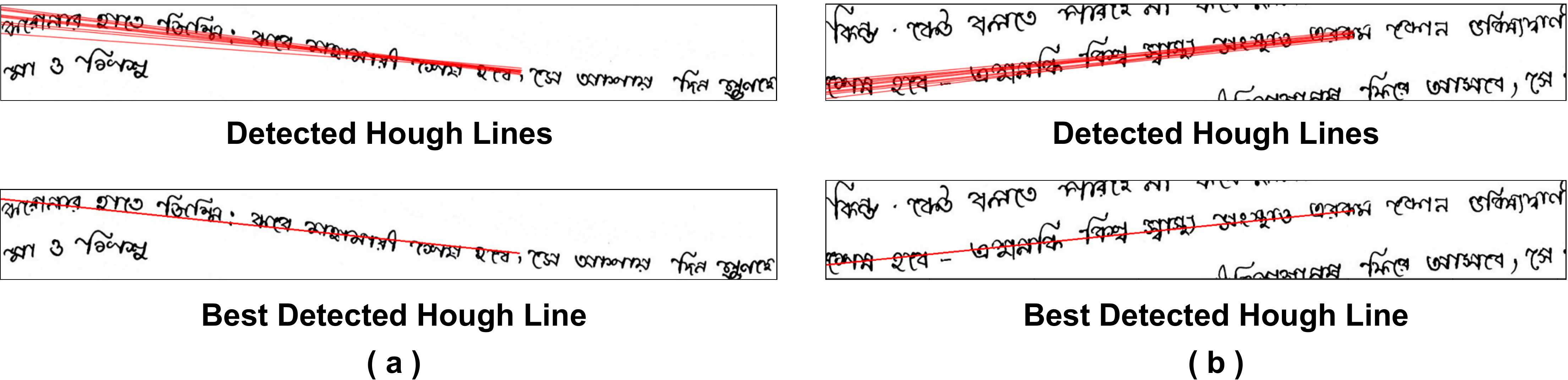}
\vspace{-6mm}
\caption{Detected (top) and the Best Detected (bottom) Hough Lines, where the main handwritten line contains, a) Positive Skew, and b) Negative Skew.} 
\label{drishti-fig5}
\vspace{-4mm}
\end{figure}

\noindent \textbf{\textit{LSkew:}} In the Bangla writings, each word consists of letters and the letters are often connected by a horizontal line called ‘mātrā’. By connecting those horizontal lines above the words using SHT, we construct straight lines, which we denote as Hough lines. Using those Hough lines, we estimate the skew angle of the main handwritten line. In terms of the representation of LSkew (Fig. \ref{drishti-fig4}), if the detected Hough lines have positive skew, the estimated skew angle will be negative; otherwise positive. We illustrate this LSkew estimation process in Fig. \ref{drishti-fig5} by taking two samples of segmented line images, where one got positive skew, and the other got negative skew. 

The SHT is applied to get the Hough lines by connecting the adjacent edge points of the main handwritten line’s words, represented in Fig. \ref{drishti-fig5} (top). Consequently, we calculated the average of all the detected Hough lines' parameters and considered this value to be the best detected Hough line. Fig. \ref{drishti-fig5} (bottom) represents the average skew angle ($\theta_{avg}$), which is the optimal skew angle of our best detected Hough line.

\begin{figure}
\centering
\vspace{-6mm}
\includegraphics[width=1.0\textwidth]{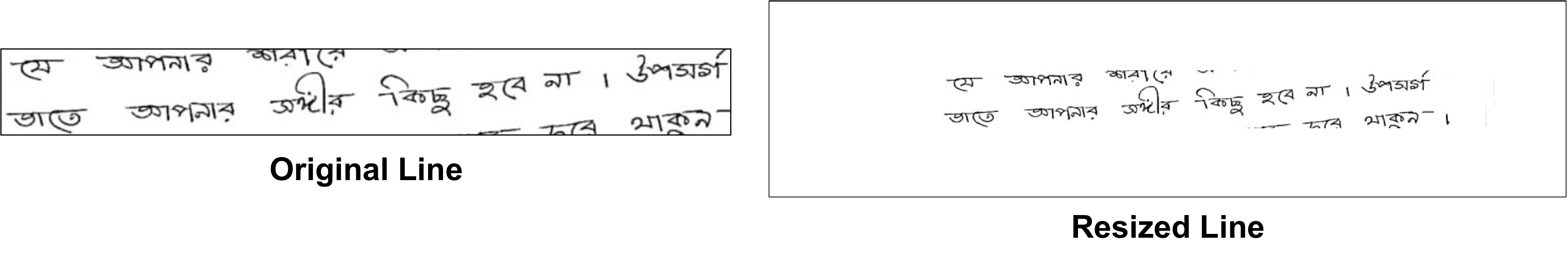}
\vspace{-6mm}
\caption{Changing the dimension of nonstandard line image before applying PHT.} 
\label{drishti-fig6}
\vspace{-4mm}
\end{figure}

\noindent \textbf{\textit{DSkew:}} In some cases, SHT fails to detect the Hough lines, despite the main handwritten line on those segmented images being well skewed. We identified that the dimension of those failed images is too small compared to the standard dimension of the line images where SHT works. Moreover, in most cases, those line images contain only a few words, in such cases, not requiring any skew correction. Therefore, we opt for the DSkew process by applying PHT. We perform up-scaling on those failed images by preserving the aspect ratio before applying PHT (shown in Fig. \ref{drishti-fig6}).

\begin{figure}
\vspace{-4mm}
\centering
\includegraphics[width=1.0\textwidth]{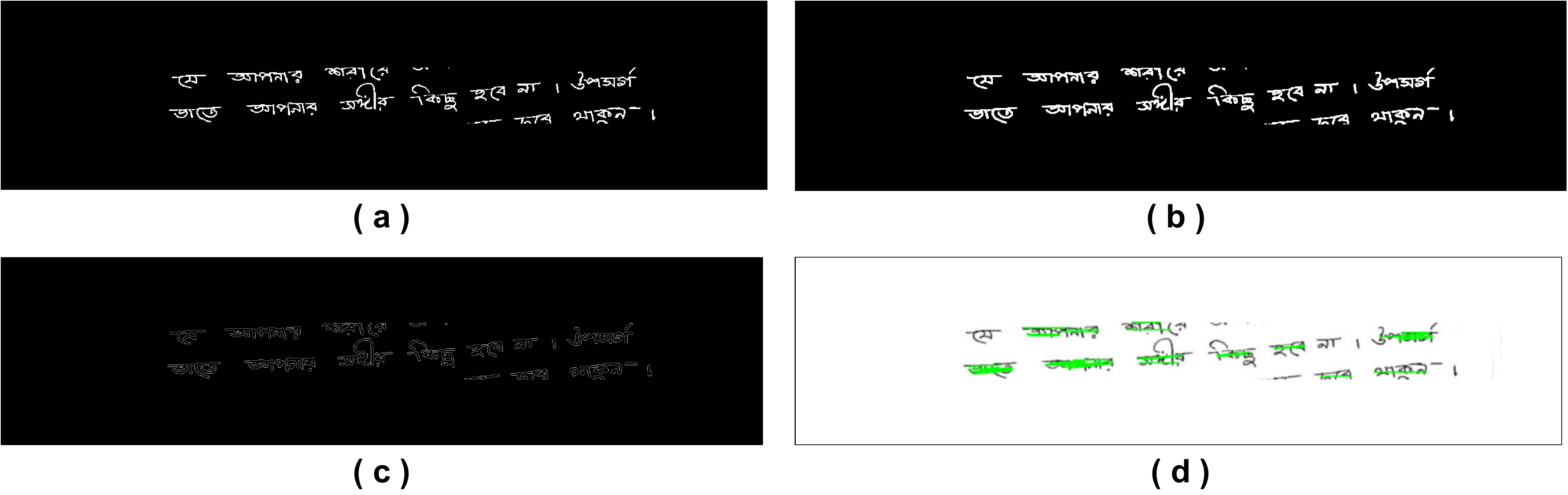}
\vspace{-6mm}
\caption{Preprocessing and Hough line detection of sample resized line image represented in Fig. \ref{drishti-fig6}; where a) Binarization, b) Morphological Dilation, c) Canny edge detection, and d) Detected Hough lines using PHT.} 
\label{drishti-fig7}
\vspace{-4mm}
\end{figure}

\noindent We apply some preprocessing steps such as image binarization and morphological operation with a 3x3 kernel to make the objects' lines and overall shape thicker and sharper. Finally, the canny edge detection method is applied before we can use the PHT. The output of preprocessing steps can be seen in Fig. \ref{drishti-fig7}.

The PHT not only joins the `matra' of words but also connects any subsets of the points of each word edges individually if there is any potential Hough line. We named it dimension skew or DSkew, as each word component in the image takes part in the skew estimation process. Like SHT, we also get the typical Hough line parameters such as ($x_1, y_1$), ($x_2, y_2$), and ($\rho, \theta$) in PHT. Hence, we applied the PHT in the edge detected image (of Fig. \ref{drishti-fig7}c) and got the Hough lines detected, shown in Fig. \ref{drishti-fig7}d. As the process detects multiple Hough lines for almost every word, therefore, each line has many $\theta$, which we denote as \textit{Degree}. To obtain the optimal skew angle of that image, we perform a voting process by dividing the $xy$ space into six cases to determine where the maximum detected Hough lines had fallen. We then take an average of those lines' parameters to find the average of Degree ($Degree_{avg}$) and consider this as the skew angle of the detected Hough lines by PHT. The six cases of the voting process and their outcomes are given in Table \ref{drishti-tab2}:

\begin{table}[h]
\vspace{-4mm}
\centering
\caption{Voting process of DSkew with their categories and outcomes.}\label{drishti-tab2}
\begin{tabular}{|c|c|c|c|}
\hline
\textbf{Based   On} & \textbf{Voting   Categories} & \textbf{Detected Hough} & \textbf{Final   outcome as an} \\
 &  & \textbf{Line Types} & \textbf{average of degrees} \\ \hline
 & $x_1$ equals $x_2$ & Vertical & Return $Degree_{avg}$  as  90° \\ \cline{2-4} 
\multirow{-2}{*}{Coordinates} & $y_1$ equals $y_2$ & Straight & Return $Degree_{avg}$ as 0° \\ \hline
 & -45°   $\le$ Degree $\le$ 0° & Positive   Skew & Return   $Degree_{avg}$ \\ \cline{2-4} 
 & -90°   $\le$ Degree \textless -45° & Negative   Skew & Return   $Degree_{avg}$ \\ \cline{2-4} 
 & 0°   \textless Degree $\le$ 45° & Negative   Skew & Return   $Degree_{avg}$ \\ \cline{2-4} 
\multirow{-4}{*}{Quadrants} & 45°   \textless Degree $\le$ 90° & Positive   Skew & Return   $Degree_{avg}$ \\ \hline
\end{tabular}
\vspace{-8mm}
\end{table}

\subsubsection{Skew Correction:}
In order to correct the estimated skew of our segmented lines, we rotate them using the Affine Transform (AT) relative to the center of the image. The rotation process for LSkew and DSkew is as follows:

\noindent\textbf{\textit{LSkew:}} After estimating the optimal skew angle ($\theta_{avg}$) using LSkew, we rotate the image with that skew angle through AT using the following two conditions:
\begin{enumerate}
  \item If the value of $\theta_{avg}$ is Negative, we rotate the image Clockwise.
  \item If the value of $\theta_{avg}$ is Positive, we rotate the image Anti-Clockwise.
\end{enumerate}

\begin{figure}
\vspace{-4mm}
\centering
\includegraphics[width=1.0\textwidth]{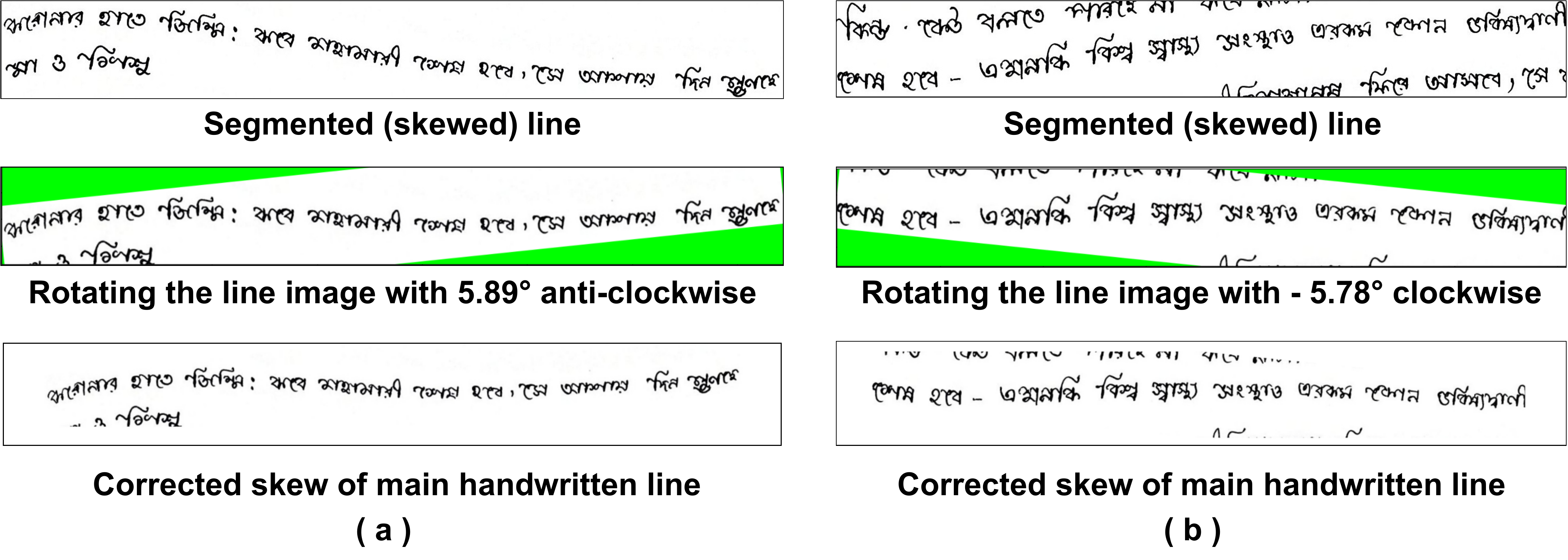}
\vspace{-8mm}
\caption{Skew correction of segmented lines using AT where the original line was, a) Negative Skewed (rotated anti-clockwise); and b) Positive Skewed (rotated clockwise).} 
\label{drishti-fig8}
\vspace{-4mm}
\end{figure}

\noindent Fig. \ref{drishti-fig8} illustrates the skew-correction for the segmented lines of Fig. \ref{drishti-fig5}.

\noindent\textbf{\textit{DSkew:}} The rotation for DSkew correction is similar to the rotation for LSkew correction, but the process of finding the optimal degree for rotation is different. Here, we calculate the optimal skew angle ($\theta_{avg}$) based on the estimated $Degree_{avg}$ from DSkew. Then according to $\theta_{avg}$, we rotate the image using AT by following the four conditions listed in Table \ref{drishti-tab3}:

\begin{table}[h]
\vspace{-4mm}
\centering
\caption{Conditions for skew correction for the process of DSkew.}\label{drishti-tab3}
% \vspace{-2mm}
% \begin{tabular}{|l|l|l|l|l|}
\begin{tabular}{|p{0.06\linewidth}|p{0.32\linewidth}|p{0.32\linewidth}|p{0.20\linewidth}|}
\hline
\textbf{No.} &  \textbf{Conditions} & \textbf{Optimal Skew ($\theta_{avg}$)} & \textbf{Rotation}\\
\hline
1. & {-45}\textdegree $\le$ $Degree_{avg}$ $\le$ {0}\textdegree & $\theta_{avg} = Degree_{avg}$ & Clockwise\\
\hline
2. & {-90}\textdegree $\le$ $Degree_{avg}$ $<$ {-45}\textdegree & $\theta_{avg} = Degree_{avg}$ + {90}\textdegree & Anti-clockwise\\
\hline
3. & {0}\textdegree $<$ $Degree_{avg}$ $\le$ {45}\textdegree & $\theta_{avg} = Degree_{avg}$ & Anti-clockwise\\
\hline
4. & {45}\textdegree $<$ $Degree_{avg}$ $\le$ {90}\textdegree & $\theta_{avg} = Degree_{avg}$ - {90}\textdegree & Clockwise\\
\hline
\end{tabular}
\vspace{-10mm}
\end{table}

\subsection{Final/Second Line Prediction and Segmentation}
\label{sec_4_4}
Final or second line prediction is applied on the skew-corrected lines to retrieve the main handwritten lines by eliminating the unwanted lines. Before that, we trim down each side of the DSkewed line image by a little portion to avoid unnecessary word prediction. Here, we consider a confidence threshold of $0.5$. We also follow a selection process when we have multiple lines even after the second line prediction, as described below:

\begin{enumerate}
  \item \textbf{The number of line predictions is one:} In this case, we segment the line with the given bounding box attributes, like in Fig. \ref{drishti-fig9}. If the width of the predicted line is less than $40\%$ of the image width, we keep it as it is.
  
    \begin{figure}[h]
        % \vspace{-6mm}
        \centering    
        \includegraphics[width=0.8\textwidth]{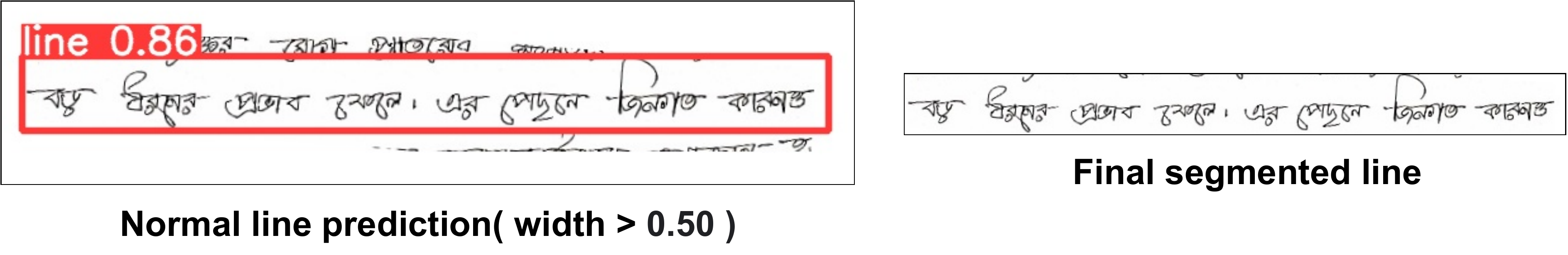}
        \vspace{-4mm}
        \caption{Line image with single line prediction and segmentation.} 
        \label{drishti-fig9}
        \vspace{-4mm}
    \end{figure}
    
  \item \textbf{The number of line predictions is two:} In this case, normally, we segment the line prediction with maximum widths, like in Fig. \ref{drishti-fig10}. But, if both the predicted line's width is less than $50\%$ of the image width, then we check their confidence and segment the line with maximum confidence. Otherwise, we keep the image as it is.
  
  \begin{figure}[h]
  \vspace{-4mm}
    \centering    \includegraphics[width=0.8\textwidth]{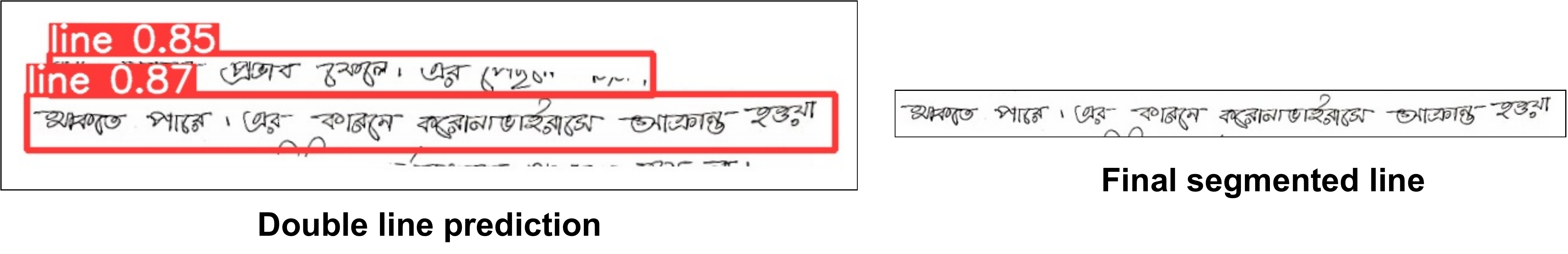}
    \vspace{-4mm}
    \caption{Line image with two line prediction and segmentation (usual case).} 
    \label{drishti-fig10}
    \vspace{-4mm}
  \end{figure}
  
  \item \textbf{The number of line predictions is three:} In this case, we segment the line which stays in the middle like in Fig. \ref{drishti-fig11}.

  \begin{figure}[h]
  \vspace{-4mm}
    \centering    
    \includegraphics[width=0.8\textwidth]{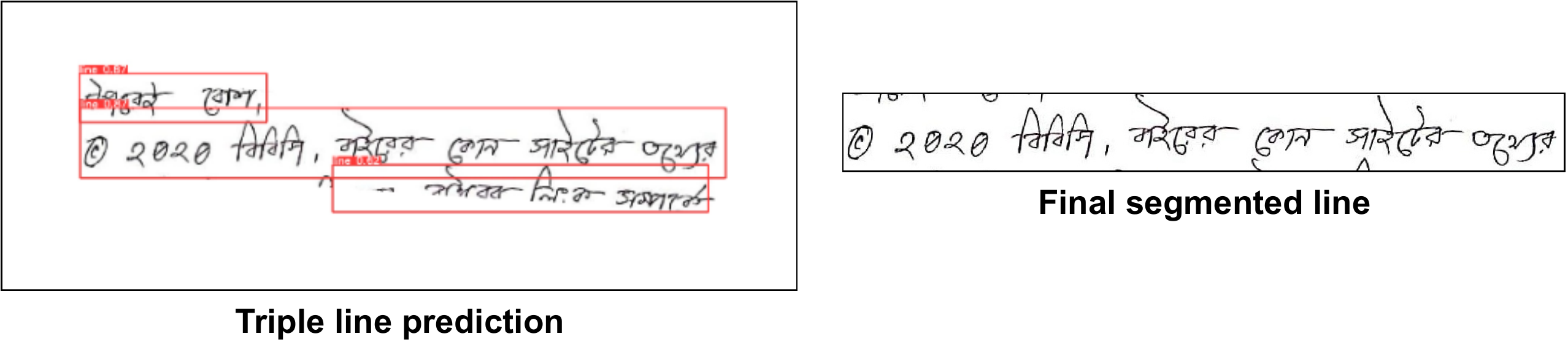}
    \vspace{-4mm}
    \caption{Line image with two line prediction and segmentation.} 
    \label{drishti-fig11}
    \vspace{-4mm}
  \end{figure}

  \item \textbf{The number of line predictions is more than three:} Unseen cases where we select and segment the line having maximum width.
\end{enumerate}

\begin{figure}[!h]
    \vspace{-4mm}
    \centering
    \includegraphics[width=1.0\textwidth]{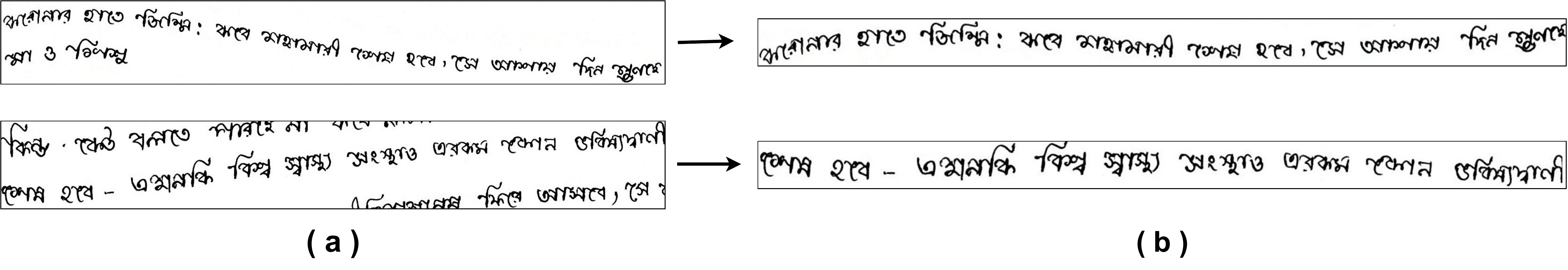}
    \vspace{-6mm}
    \caption{a) Initial line segmentation by YOLO containing mostly curvilinear or skewed handwritten lines with noises, and b) Final segmented lines by our line segmentation approach, which are straight and without any unnecessary lines.} 
    \label{drishti-fig12}
    % \vspace{-2mm}
\end{figure}

\noindent As the segmented lines have passed through the pre-processing, rotation, and final line segmentation process, we now have our final lines segmented from the handwritten document images. Note that, we also keep track of the predicted \textit{line numbers} within the document for future recognition purposes. Fig. \ref{drishti-fig12} illustrates the resultant final line segmentation of the lines represented in Fig. \ref{drishti-fig5}.

\subsection{Word Prediction and Segmentation}
\label{sec_4_5}
\begin{figure}
    \vspace{-4mm}
    \centering
    \includegraphics[width=1.0\textwidth]{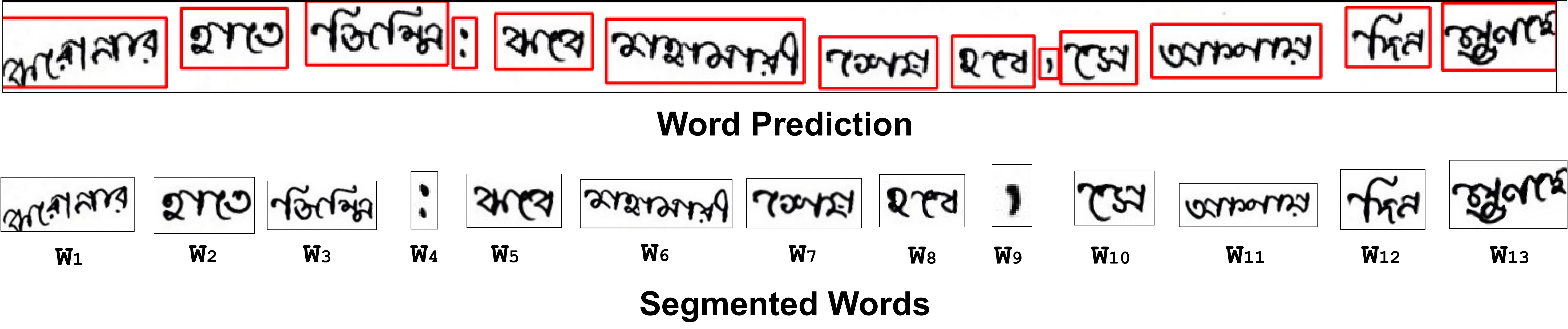}
    \vspace{-6mm}
    \caption{Word prediction and segmentation on skew corrected final segmented lines; where \textbf{$W_i$} is the $i^{th}$ word within the line.} 
    \label{drishti-fig13}
    \vspace{-10mm}
\end{figure}

We perform word prediction on the Final segmented lines by directly employing our custom YOLO word model, where we set the confidence threshold to be $0.4$. We also sort the predictions based on the horizontal axis of the lines in order to get the position of a particular word in that line for future recognition purposes. Fig. \ref{drishti-fig13} illustrates word prediction and segmentation from the running example.

\section{Experimental Results}
\vspace{-2mm}
In this section, we evaluate the efficiency of our line and word segmentation approach on the \texttt{BN-HTRd} dataset. We will also compare our results with an unsupervised line segmentation approach of \texttt{BN-HTR\_LS} system \cite{rahman2023bn}. 

\vspace{-2mm}
\subsection{Evaluation Matrices}
Two bounding boxes (lines) are considered a one-to-one match if the total matching pixels exceed or equal the evaluator's approved threshold ($T_a$). Let \textit{N} be the number of ground-truth elements, \textit{M} be the count of detected components, and \textit{o2o} be the number of one-to-one matches between \textit{N} and \textit{M}; the Detection Rate (DR) and Recognition Accuracy (RA) are equivalent of Recall and Precision. Combining these, we can get the final performance metric FM (similar to F-score) using the equation below:

\begin{equation} \label{eqn_5_1}
DR = \frac{o2o}{N},\quad RA =\frac{o2o}{M},\quad FM = \frac{2DR*RA}{DR + RA}
\end{equation}

\vspace{-2mm}
\subsection{Line Segmentation}

For the evaluation of our \texttt{BN-DRISHTI} line segmentation approach, we first did the \textbf{Quantitative analysis} on the test set of 75 handwritten document images from the BN-HTRd dataset containing 1397 (\textit{N}) manually annotated ground truth lines. Our segmentation approach's final line prediction was 1396 (\textit{M}). Among those, the number of \textit{o2o} matches was 1314. However, by using only YOLO trained model, we got 1433 (\textit{M}) which implies that YOLO predicted 37 more redundant lines as compared to our approach, making our approach much superior. These results are listed in rows 2-3 of Table \ref{drishti-tab4}.

\begin{figure}[h]
\vspace{-4mm}
    \centering
    \includegraphics[width=0.6\textwidth]{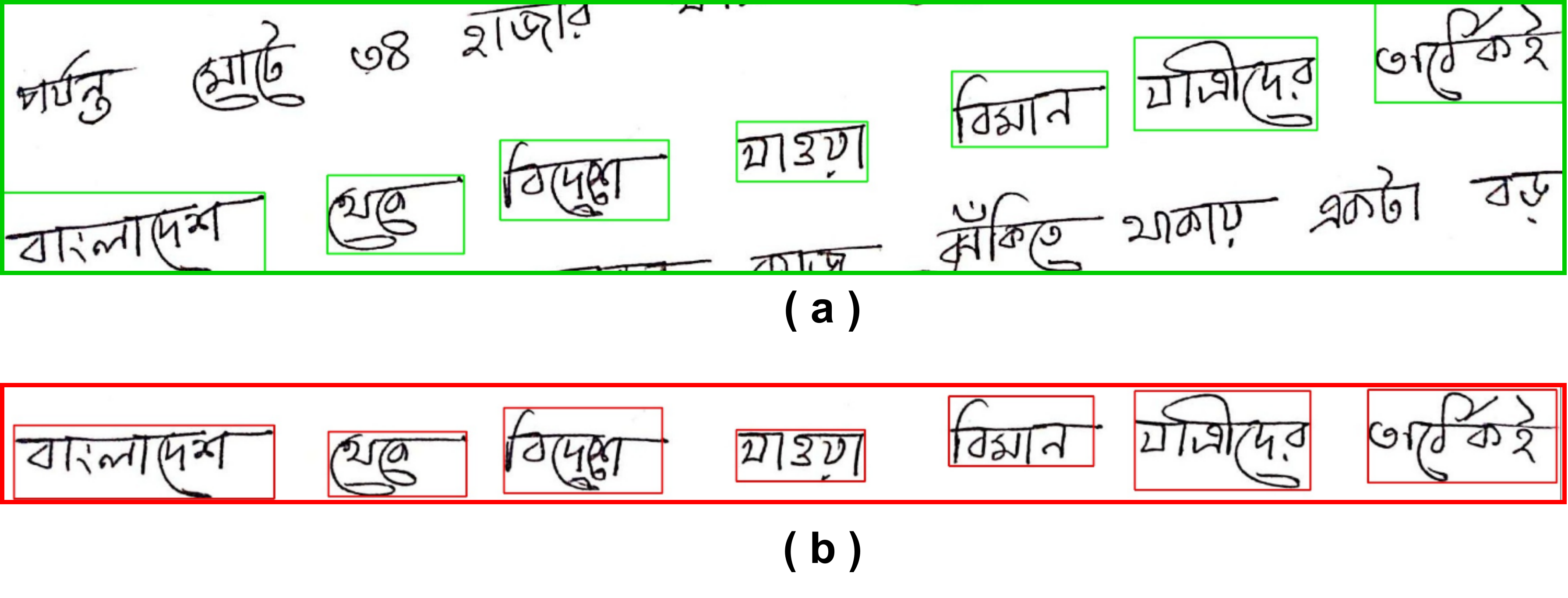}
    \vspace{-4mm}
    \caption{a) Ground truth annotation on skewed line; Vs. b) Prediction on straight line.}
    \label{drishti-fig14}
    \vspace{-5mm}
\end{figure}

\noindent After analyzing the line’s ground truth and prediction bounding boxes visually (see Fig. \ref{drishti-fig14}), we came to the conclusion that the overlap between them for each line is not quite accurate since we performed skew correction before segmenting the line images. Thus, in automatic or quantitative evaluation the results we are getting are not as significant as we were expecting, since almost every line of the document images was segmented perfectly. Hence, we decided to do a \textbf{Qualitative evaluation} by going through all the ground truth and predictions manually to find the \textit{o2o} for each handwritten document. And the overall \textit{o2o} match was 1396, which is equal to the final line predictions we were getting. In Table \ref{drishti-tab4} we put together the relative performance of our line segmentation approach’s (\texttt{BN-DRISHTI}) quantitative and qualitative analysis as compared to the unsupervised approach of \texttt{BN-HTR\_LS} system\footnote{\textbf{BN-HTR\_LS Codebase:} \url{https://github.com/shaoncsecu/BN-HTR\_LS}} \cite{rahman2023bn} where they only performed line segmentation on the same dataset.

\begin{table}
\vspace{-4mm}
\centering
\caption{Comparison of line segmentation results on BN-HTRd test sets.}\label{drishti-tab4}
\begin{tabular}{|p{0.4\linewidth}|p{0.08\linewidth}|p{0.08\linewidth}|p{0.08\linewidth}|p{0.1\linewidth}|p{0.1\linewidth}|p{0.1\linewidth}|p{0.1\linewidth}|}
\hline
\textbf{Approaches} & \textbf{N}	&  \textbf{M} & \textbf{o2o} & \textbf{DR($\%$)} & \textbf{RA($\%$)} & \textbf{FM($\%$)} \\
\hline
BN-HTR\_LS \cite{rahman2023bn} & 2915 & 3437 & 2591 & 88.88 & 75.38 & 81.57 \\
\hline
YOLO line model & 1397 & 1433 & 1314 & 94.06 & 91.7 & 92.86 \\
\hline
\textbf{BN-DRISHTI} (Quantitative) & 1397 & 1396 & 1314 & 94.06 & 94.13 & 94.09 \\
\hline
\textbf{BN-DRISHTI} (Qualitative) & 1397 & 1396 & 1396 & \textbf{99.93} & \textbf{1.00} & \textbf{99.97} \\
\hline
\end{tabular}
\vspace{-10mm}
\end{table}

% \noindent In Table \ref{drishti-tab4}, it is clearly visible that our BN-HTR\_LWS’s line segmentation approaches result of qualitative analysis is almost near perfect and also far beyond the other analysis on the table. And that truly justifies our systems line segmentation performance.

% In Fig. \ref{fig24}, we also illustrated the F1 vs Confidence curve on the test set (see Table \ref{drishti-tab1}) where F1 hits 0.86 in (a); and the Precision vs Recall curve where the mean average precision (mAP) hits 0.94 in (b) at 0.8 threshold.

% \begin{figure}
%     \centering
%     % \includegraphics[width=1.0\textwidth]{images/24.png}
%     \includegraphics[width=1.0\textwidth]{images/performance_curve_line_seg.pdf}
%     % \vspace{-4mm}
%     \caption{Performance of our line segmentation approach on the test data, where a) F1 vs. Confidence, and b) Precision vs. Recall.} 
%     \label{fig24}
% \end{figure}

\subsection{Word Segmentation} 
For this experiment, we used 10,414 manually annotated ground truth words within the line images of the test set's 75 handwritten documents. Our word model predicted 10,348 words. Table \ref{drishti-tab5} shows the score of \textbf{Quantitative} analysis.

\begin{table}[h]
\vspace{-4mm}
\centering
\caption{Quantitative evaluation of our word segmentation on BN-HTRd test sets.}\label{drishti-tab5}
\vspace{-1mm}
\begin{tabular}{|p{0.21\linewidth}|p{0.15\linewidth}|p{0.11\linewidth}| p{0.11\linewidth}| p{0.11\linewidth}|}
\hline
\textbf{Ground Truths} & \textbf{Prediction} & \textbf{DR ($\%$)} & \textbf{RA ($\%$)} & \textbf{FM ($\%$)}\\
\hline
10,414 & 10,348 & 15.2 & 17.7 & 16.0\\
\hline
\end{tabular}
\vspace{-5mm}
\end{table}

% \begin{table}[h]
% \vspace{-4mm}
% \centering
% \caption{Quantitative evaluation of word segmentation; where DR, RA, and FM represent Recall, Precision and F1-score, respectively.}\label{drishti-tab5}
% \begin{tabular}{|p{0.17\linewidth}|p{0.14\linewidth}|p{0.08\linewidth}| p{0.08\linewidth}| p{0.08\linewidth}|p{0.09\linewidth}|p{0.13\linewidth}|}
% \hline
% \textbf{Ground Truths} & \textbf{Prediction} & \textbf{DR ($\%$)} & \textbf{RA ($\%$)} & \textbf{FM ($\%$)} & \textbf{mAP (0.8)} & \textbf{mAP (0.8:0.95)}\\
% \hline
% 10,414 & 10,348 & 15.2 & 17.7 & 16.0 & 0.103 & 0.0381\\
% \hline
% \end{tabular}
% \vspace{-4mm}
% \end{table}

\noindent Again for the same aforementioned reason, the quantitative evaluation does not do justice to our approach's true word segmentation capabilities. Hence, we visually compared the ground truths against our predictions and found that the position of the words bounding box has changed drastically due to the changes in image dimension during our line segmentation approach, as illustrated in Fig. \ref{drishti-fig14}. This occurred because the original ground truth annotation was on the skewed lines, and our word prediction was done on the skew-corrected straight lines. Thus, after analyzing the ground truth and prediction bounding boxes, we came to the conclusion that the evaluation will not be fair if done automatically. Therefore, we again opt for a manual \textbf{Qualitative} analysis. We show both the quantitative and qualitative results in Table \ref{drishti-tab6}.

\begin{table}[h]
\vspace{-4mm}
\centering
\caption{Results of our word segmentation approach on the original ground truth (skewed) vs. skew-corrected (straight) lines from the BN-HTRd test sets.}\label{drishti-tab6}
\begin{tabular}
% \begin{tabular}{|c|c|c|c|c|c|c|c|}
{|l|l|p{0.09\linewidth}|p{0.09\linewidth}|p{0.07\linewidth}|p{0.07\linewidth}|p{0.07\linewidth}|}
\hline
\textbf{Analysis} & \textbf{Word prediction on} & \textbf{N} & \textbf{M} & \textbf{DR} & \textbf{RA} & \textbf{FM} \\ \hline
Quantitative & First segmented (Skewed) line & 10,414 & 10,383 & 0.39 & 0.45 & 0.42 \\
\hline
Quantitative & Final segmented (Straight) line & 10,414 & 10,348 & 0.15 & 0.17 & 0.16 \\ 
\hline
Qualitative & Final segmented (Straight) line & 10,414 & 10,348 & 0.98 & 0.98 & \textbf{0.98 }\\ 
\hline
\end{tabular}
\vspace{-4mm}
\end{table}

\noindent In Table \ref{drishti-tab6}, the qualitative analysis results perfectly justify our systems word segmentation capabilities. We also emphasize that word segmentation is far more precise when combined with our skew correction strategy.

\subsection{Automatic Annotation}
We used the extra 87 folders containing 805 document images without ground truths to get an idea about the efficiency of our system's automatic line and word annotation capability. We randomly picked 151 document images and their predictions and compared them manually. The obtained manual evaluation scores are given in Table \ref{drishti-tab7}, and the results far exceed our expectations.

\begin{table}[h]
\vspace{-4mm}
\centering
\caption{Results of automatic annotation on the unannotated portion of BN-HTRd.}\label{drishti-tab7}
\vspace{-1mm}
\begin{tabular}{|p{0.1\linewidth}|p{0.20\linewidth}|p{0.15\linewidth}|p{0.09\linewidth}|p{0.11\linewidth}|p{0.11\linewidth}|p{0.11\linewidth}|}
\hline
\textbf{Class} & \textbf{Ground Truth} & \textbf{Prediction} & \textbf{o2o} & \textbf{DR ($\%$)} & \textbf{RA ($\%$)} & \textbf{FM ($\%$)}\\
\hline
Line &	2829 &	2840 &	2829 &	100 &	99.61 &	99.80\\
\hline
Word &	19,402 &	19,393  &	19,299 &	99.4 &	99.51 &	99.49\\
\hline
\end{tabular}
\vspace{-6mm}
\end{table}

\vspace{-2mm}
\subsection{Comparative Analysis}

\textbf{ICDAR 2013 Dataset \cite{stamatopoulos2013icdar}}: This handwriting segmentation contests dataset contains 50 images for Bangla. As ground truth ($N$), we got 879 lines and 6,711 words; against which our system segmented 874 lines and 6,667 words ($M$). We choose team Golestan-a, Golestan-b, and INMC for performance comparison, as the Golestan method outperforms all other contestants with an overall score (SM) of $94.17\%$. And for Line segmentation, the INMC method was on the top with a $98.66\%$ FM score. The comparison in Table \ref{drishti-tab8} indicates that our system outperforms Golestan and INMC team's SM scores by a good margin. While our word segmentation results absolutely smashed the competitors, the line segmentation score was only second to INMC by a narrow margin.

% \usepackage{multirow}
\begin{table}[h]
\vspace{-4mm}
\centering
\caption{Comparison among top teams of ICDAR 2013 and our BN-DRISHTI system.}\label{drishti-tab8}
\vspace{-1mm}
\begin{tabular}{|c|c|c|c|c|c|c|c|c|}
\hline
\textbf{Systems} & \textbf{Class} & \textbf{N} & \textbf{M} & \textbf{o2o} & \textbf{DR (\%)} & \textbf{RA (\%)} & \textbf{FM (\%)} & \textbf{SM (\%)} \\ \hline
\multirow{2}{*}{Golestan-a} & Lines & 2649 & 2646 & 2602 & 98.23 & 98.34 & 98.28 & \multirow{2}{*}{94.17} \\ \cline{2-8}
 & Words & 23525 & 23322 & 21093 & 89.66 & 90.44 & 90.05 &  \\ \hline
\multirow{2}{*}{Golestan-b} & Lines & 2649 & 2646 & 2602 & 98.23 & 98.34 & 98.23 & \multirow{2}{*}{90.06} \\ \cline{2-8}
 & Words & 23525 & 23400 & 21077 & 89.59 & 90.07 & 89.83 &  \\ \hline
\multirow{2}{*}{INMC} & Lines & 2649 & 2650 & 2614 & 98.68 & 98.64 & \textbf{98.66} & \multirow{2}{*}{93.96} \\ \cline{2-8}
 & Words & 23525 & 22957 & 20745 & 88.18 & 90.36 & 89.26 &  \\ \hline
\multirow{2}{*}{\textbf{BN-DRISHTI}} & Lines & 879 & 874 & 863 & 98.18 & 98.74 & 98.46 & \multirow{2}{*}{\textbf{96.65}} \\ \cline{2-8}
 & Words & 6711 & 6677 & 6348 & 98.74 & 95.07 & \textbf{94.83} &  \\ \hline
\end{tabular}
\vspace{-4mm}
\end{table}

\newpage
\noindent \textbf{BanglaWriting Dataset \cite{mridha2021banglawriting}}: It comprises 260 full-page Bangla handwritten documents and only the words ground truth. We manually evaluated the word segmentation results using randomly selected 50 document images from this dataset, as the word annotation was done directly over the document without any intermediate line annotation. Those selected 50 images contain 4409 words, and our system correctly segmented 4186 words against them. Table \ref{drishti-tab9} indicates how our system performed on the BanglaWriting dataset.

\begin{table}
\vspace{-4mm}
\centering
\caption{Word segmentation results on fifty images of BanglaWriting dataset.}\label{drishti-tab9}
\vspace{-2mm}
\begin{tabular}{|p{0.24\linewidth}|p{0.09\linewidth}|p{0.09\linewidth}|p{0.09\linewidth}|p{0.12\linewidth}|p{0.12\linewidth}|p{0.12\linewidth}|}
\hline
\textbf{Task} & \textbf{N} & \textbf{M} & \textbf{o2o} & \textbf{DR ($\%$)} & \textbf{RA ($\%$)} & \textbf{FM ($\%$)}\\
\hline
Word Segmentation &	4409 &	4219 &	4186 &	94.9 &	99.2 &	97.0\\
\hline
\end{tabular}
\vspace{-4mm}
\end{table}

\noindent \textbf{WBSUBNdb\_text Dataset \cite{halder2018content}}: This publicly available dataset has been used by two of the most prominent line \cite{rakshit2023generalized} and word \cite{agarwal2022word} segmentation methods for evaluation. As it contains 1352 Bangla handwriting without any ground truth, we only performed a qualitative analysis similar to the settings mentioned in those papers. We positioned our approach against these systems in Table \ref{drishti-tab10}.   

% \usepackage{multirow}
\begin{table}[h]
\vspace{-4mm}
\centering
\caption{Comparison of segmentation results based on WBSUBNdb\_text dataset.}\label{drishti-tab10}
\vspace{-1mm}
\begin{tabular}
{|p{0.20\linewidth}|p{0.15\linewidth}|p{0.15\linewidth}|p{0.15\linewidth}|p{0.15\linewidth}|}
\hline
\textbf{Systems} & \textbf{Class} & \textbf{DR (\%)} & \textbf{RA (\%)} & \textbf{FM (\%)} \\
\hline
\multirow{2}{*}{WBSUBNdb} & Lines \cite{rakshit2023generalized} & 96.99 & 97.07 & 97.02 \\ \cline{2-5}
 & Words \cite{agarwal2022word} & 86.96 & 93.25 & 90.0 \\
\hline
\multirow{2}{*}{\textbf{BN-DRISHTI}} & Lines & 99.27 & 99.44 & \textbf{99.35} \\\cline{2-5}
 & Words & 96.85 & 97.18 & \textbf{97.01} \\
 \hline
\end{tabular}
\vspace{-8mm}
\end{table}

\section{Conclusions}
\vspace{-1mm}
The main contribution of this research is the significant improvement in line and word segmentation for Bangla handwritten scripts, which lays the foundation of our envisioned Bangla Handwritten Text Recognition (HTR). To alleviate the shortage of Bangle document-level handwritten datasets for future researchers, we have extended our BN-HTRd dataset. Currently, it is the largest dataset of its type with line and word-level annotation. Moreover, keeping the recognition task in mind, we have stored the words' Unicode representation against their position in the ground truth text. The main recipe behind our approach's overwhelming success is a two-layer line segmentation technique combined with an intricate skew correction in the middle. Our proposed line segmentation approach has achieved a near-perfect benchmark evaluation score in terms of F measure ($99.97\%$) compared to the unsupervised approach ($81.57\%$) of BN-HTR\_LS \cite{rahman2023bn}. The word segmentation technique also achieved an impressive score ($98\%$) on the skew-corrected lines by our system compared to the skewed lines. Furthermore, we have compared our method against the previous SOTA systems on three of the most prominent Bangla handwriting datasets. Our approach outperformed all those methods by a significant margin, making our \texttt{BN-DRISHTI} system a new state-of-the-art for Bangla handwritten segmentation task. We aim to expand our work by integrating supervised word recognition to build an “End-To-End Bangla Handwritten Image Recognition system”.

\bibliographystyle{splncs04}
\bibliography{references}
\end{document}